\documentstyle[ijcai95,epsf]{article}

\title{Ubiquitous Talker:\\
       Spoken Language Interaction with Real World Objects}

\author{Katashi Nagao {\rm and} Jun Rekimoto\\
Sony Computer Science Laboratory Inc.\\
3--14--13 Higashi-gotanda,
Shinagawa--ku, Tokyo 141, Japan\\
E-mail: \{nagao,rekimoto\}@csl.sony.co.jp}

\begin{document}

\maketitle

\begin{abstract}
{\em Augmented reality} is a research area that tries to embody an electronic
information space within the real world, through computational devices.
A crucial issue within this area, is the recognition of real world objects
or situations.
In natural language processing, it is much easier to determine
interpretations of utterances, even if they are ill-formed, when the
context or situation is fixed. We therefore introduce robust, natural
language processing into a system of augmented reality with situation
awareness. Based on this idea, we have developed a portable system, called
the {\em Ubiquitous Talker}.  This consists of an LCD display that reflects
the scene at which a user is looking as if it is a transparent glass,
a CCD camera for recognizing real world objects with color-bar ID codes,
a microphone for recognizing a human voice and a speaker which outputs a
synthesized voice. The Ubiquitous Talker provides its user with some
information related to a recognized object, by using the display and
voice. It also accepts requests or questions as voice inputs. The user
feels as if he/she is talking with the object itself through the system.
\end{abstract}

\section{Introduction}

There are many situations where we want to interact with the surrounding
real world. We would also like to communicate with the objects used in
our everyday life. {\em Augmented reality} is a research area that tries to
incorporate an electronic information space into the real world, by
means of computational devices\footnote{The term `augmented reality'
usually refers to a variant of virtual reality, that uses see-through
head-mounted displays to overlay computer-generated images on the user's
real world view. For example, see \cite{Feiner93}. However, we use it here
with a more general meaning.}.
This approach enriches, rather than replaces the real world (i.e., a virtual
reality), by providing valuable information, such as descriptions of
objects, navigational help in places, and instructions for performing
physical tasks.
Augmented reality essentially requires the ability to recognize real world
objects/situations. There are several approaches to situation awareness,
such as detection of physical objects using visual processing,
detection of location/orientation by positioning systems, and
communication with physically embedded computers (i.e., ubiquitous
computing \cite{Weiser93}). For situation awareness, we employ a
colored {\em barcode} system \cite{Rekimoto94}. In this system, any
real world object has a color-bar tag attached to it that makes it
easily identifiable.

On the other hand, in order to make natural language processing,
especially spoken language processing, more practical, we must restrict
or constrain the domains, contexts, or tasks, since it requires a
potentially broad search space on a phonetic and linguistic level.
Recently, there has been a big trend in multimodal approaches to combine
verbal and nonverbal modalities in human-computer communication.
Various sorts of nonverbal information play a role in setting the
situational context, which is useful in restricting the hypothesis
space constructed during language processing.
When a context or a situation is fixed by using nonverbal information,
the interpretation of utterances becomes much easier, even
if the utterances are ill-formed.
In other words, the correct interpretation of natural language utterances
essentially requires the integration of both linguistic and non-linguistic
contexts. Understanding multimodal dialogues is not possible without
some account of the role of the non-linguistic context. Considering such a
context, results in knowledge bases that are very efficient and robust.
We have therefore introduced robust natural language processing into a
system of augmented reality.

We have developed a portable system, called the {\em Ubiquitous Talker}.
This consists of an LCD display which presents the view at which a user is
looking, as if it is a transparent glass, a CCD camera for recognizing
real world objects with color-bar ID codes, a microphone for
recognizing a human voice, and a speaker that outputs a synthesized
voice. The Ubiquitous Talker augments reality with some additional
information related to a recognized object/situation. Such information
is conveyed by using the LCD display and voice. The system accepts and
interprets user voice requests and questions. The user may feel as if
he/she is talking with the object itself through the system.

In the rest of this paper, we discuss a combination of verbal and nonverbal
modalities and its role in effective interaction, explain some
implementation issues and sample applications of the Ubiquitous Talker,
and compare this research with some related work.

\section{Situated Interaction}

A real world situation includes a place where a human is, a time when an event
occurs, living and non-living things that exist in the vacinity, and a
physical action that he or she does (e.g., looking at something).

Using situation awareness, humans can naturally interact with the system
by spoken language without being specially conscious of domains or regulations
that constrain the system. This type of interaction, is called {\em situated
interaction} and is very useful in certain situations. In this case,
language use can be flexible and robust, since a shared situation apparently
reveals a topic or focus of the dialogue and objects referred to by deictic
expressions.
By recognizing situations and knowing behaviors that humans usually do in those
situations, the system can be aware of humans' intentions and predict what they
do next. Also the system can clarify humans' desires by accepting information
conveyed through voices and/or actions.

The semantics of natural language expressions is anchored to real world
objects and events by means of pointing, demonstrating actions and deictic
expressions such as ``this,'' ``that,'' ``here,'' ``there,'' ``then,'' and
``now.'' Some research on dialogue systems has combined deictic gestures
and natural language such as Put-That-There \cite{Bolt80},
CUBRICON \cite{Neal88}, and {\sc AlFresco} \cite{Stock91}.

The efficiency of situated conversation causes ambiguities of conversational
contents. Ambiguities are inherent in natural language communication; however,
they can be resolved more easily when the real world situation is recognized,
than resolved by verbally-conveyed information alone.

In addition, the focus of attention or the focal point plays a very important
role in processing applications within a broad hypothesis space, such as
speech recognition.
One example of focusing modality is following the humans' looking behavior.
Fixation or gaze is useful for the dialogue system to determine the context
of the humans' interests. For example, when a person is looking at a car,
what he/she says at that time may be related to the car.

Prosodic information (e.g., voice tones) in the humans' utterances
also helps to determine focus. For example, if a person puts a stress on a
particular word, he/she has a special interest in an object or an event
connected to the word.

Combining gestural information with spoken language comprehension
shows another example of how context may be determined by the user's
nonverbal behavior \cite{Oviatt93}. This research uses multimodal
forms that prompt a user to speak or write into labeled fields. The
forms are capable of guiding and segmenting inputs, of conveying the
kind of information the system is expecting, and of reducing
ambiguities in utterances by restricting syntactic and semantic
complexities.

On the other hand, humans can move from one situation to another by physical
actions (e.g., walking). When moving closer toward a situation, he/she gets
information related to the situation that is confronting him/her. This
can be an intuitive process for information seeking.
Walking through real world situations is a more natural way of information
retrieval than a search within a complex information space.
Our method can be considered as putting a retrieval cue to a situation.
For example, if a person wants to read a book, he/she naturally has the idea
of going to a place where a bookshelf exists. This means that a situation
that includes a bookshelf can be a retrieval cue for searching for books.

Human memories consist of mixtures of real world situations and
information that was accessed in those situations.  Therefore, recognizing
a real world situation can be a trigger for extracting a memory partially 
matched with that situation and associating information related to the memory.

In the next section, we present a prototype system based on the idea of
situated interaction. The system recognizes real world situations/objects
by putting ID tags on objects, then performs situated conversation. 
An important point is that the system accepts the real world situation
as a new input modality and integrates it into spoken dialogue processing.

\section{Ubiquitous Talker}

The Ubiquitous Talker is a speech dialogue system with situation
awareness of the real world. When it detects a real world object,
interactions with it makes its user feel as if he/she is talking with
the focused object itself.
This is a (pseudo) portable system like so-called PDAs (Personal Digital
Assistants)\footnote{In fact, only the LCD/Camera unit is portable, and it is
connected with a workstation. In the near future, this connection will be
wireless.}.

\subsection{System Configuration}

\begin{figure*}[htbp]
\centering{\epsfile{file=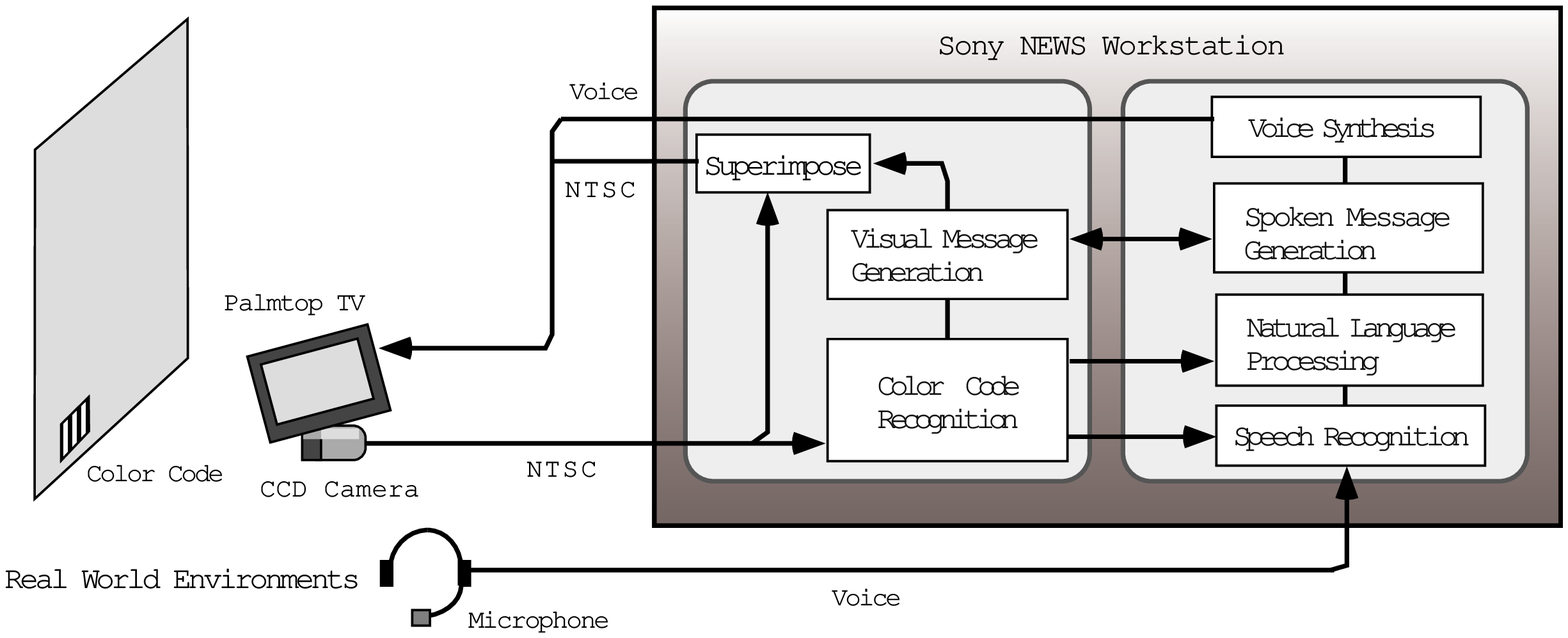,scale=.65}}
\caption{System Overview of the Ubiquitous Talker}
\label{UbiTalk}
\end{figure*}

Figure \ref{UbiTalk} shows the system configuration.

\begin{figure}[htbp]
\centering{\epsfile{file=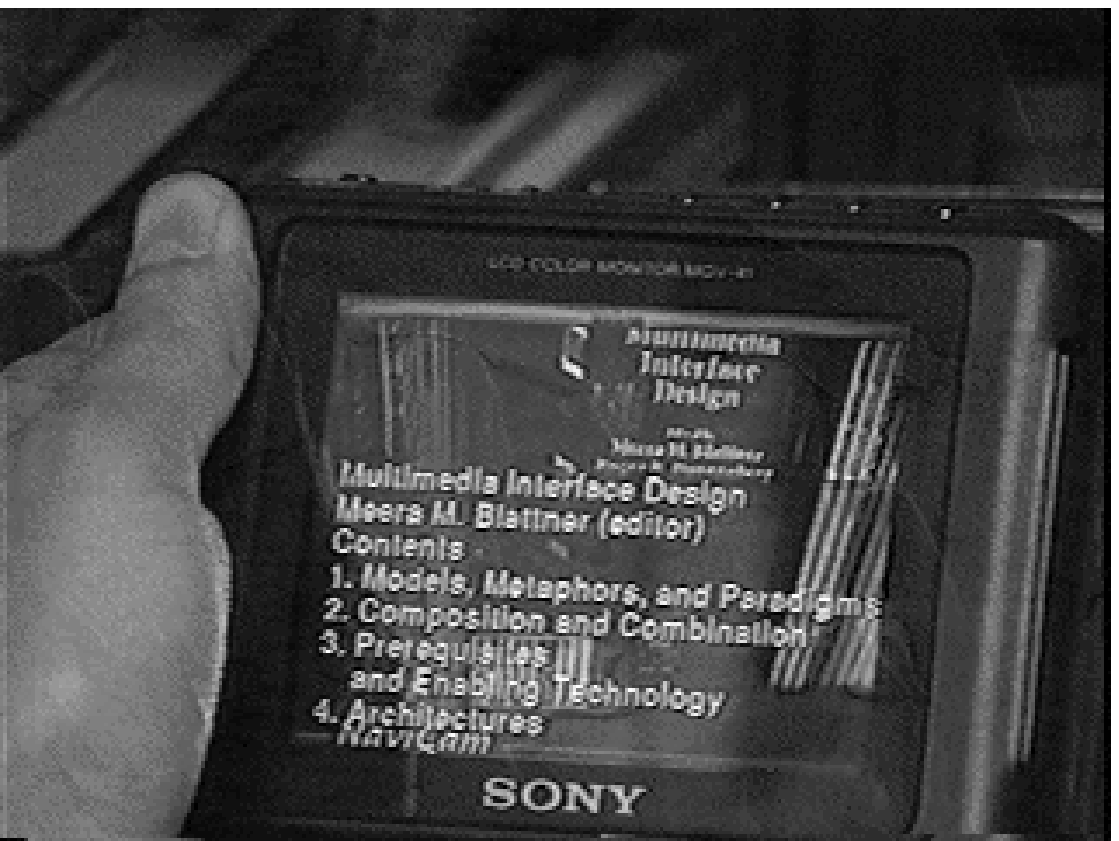,scale=.75}}
\caption{LCD Unit of the Ubiquitous Talker}
\label{NaviCam}
\end{figure}

This system basically consists of two subsystems. One subsystem recognizes
a number of real world situations that include objects with color-bar
ID codes, and shows some textual and graphical information
superimposed on an LCD display as shown in Figure \ref{NaviCam}\footnote{
This subsystem can be used independently of the speech dialogue subsystem,
which is called {\em NaviCam} (NAVIgation CAMera) \cite{Rekimoto95a}}.

The other subsystem recognizes and interprets user speech inputs and
generates voice outputs.
These two subsystems communicate with each other. The image
(color-code) recognizer triggers the speech recognizer and sends a
message to it in order to select the appropriate vocabulary and
grammar for analyzing the spoken utterances. It also selects a knowledge
base for processing the utterances and generating voice responses. The
spoken message generator and the visual message (text and graphics)
generator also communicate with each other and synchronize the time
at which to say/show information.
The user can verbally select an item from the displayed menu, or ask some
questions according to guidance conveyed by the voice and/or the text.

When the system recognizes a real world object, for instance, a
calendar on the wall, the system sends a message such as ``Today is
April 24, 1995. Your schedule is,'' and displays a timetable of the
user's schedule.
Then, the user asks ``What about tomorrow?,'' and the system replies
``Your schedule tomorrow is'' and proceeds as before.

\begin{figure}[htbp]
\centering{\epsfile{file=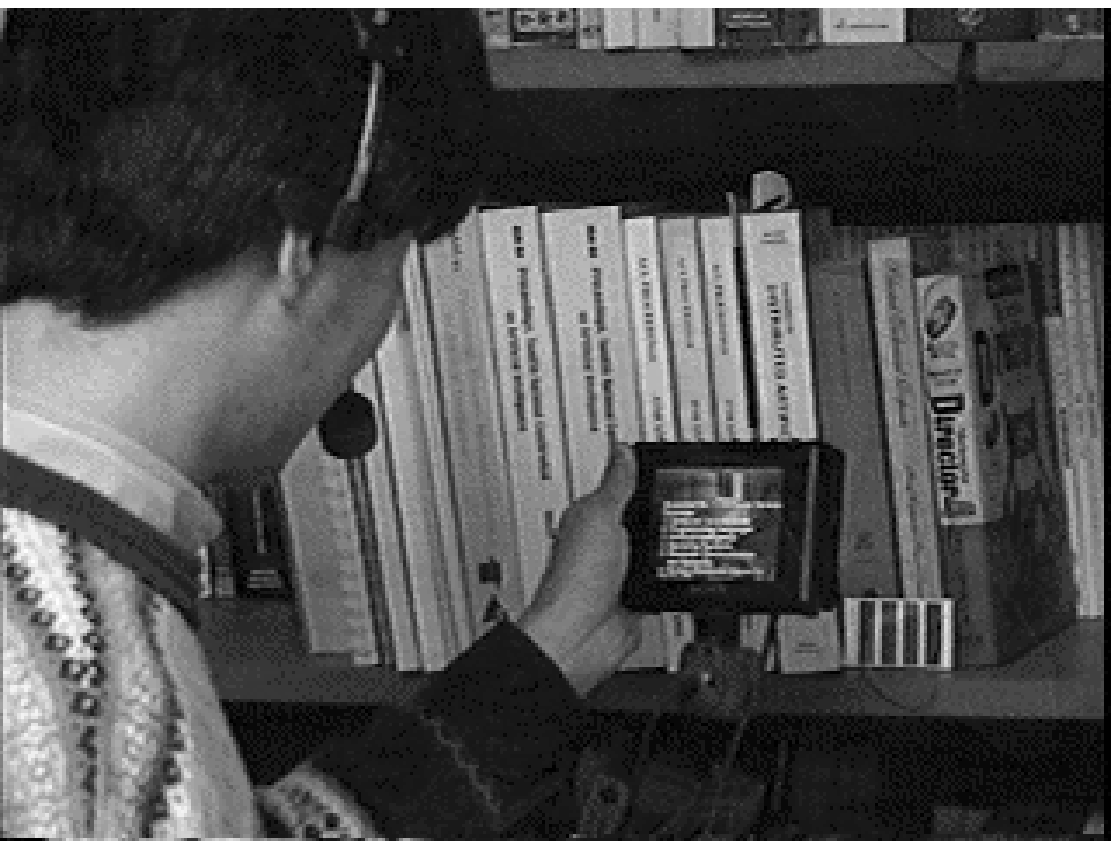,scale=.75}}
\caption{Ubiquitous Talker in Use}
\label{Snapshot}
\end{figure}

Figure \ref{Snapshot} shows a snapshot of conversation with an object
through the system.

\subsection{Situation Awareness}

To easily recognize real world situations/objects, we use a colored
{\em barcode} system as a real world identification \cite{Rekimoto94}.
The color-code consists of a number of blue and red stripes that encodes
the ID of a real world object.

An image input from a small CCD camera is processed in real time by a
workstation.
The image recognizer continuously scans any objects with the color-code.
The system generates a synthesized image by superimposing visual messages
related to the color-code on the real world image obtained
from the camera. The output image is shown on the LCD display.
Image processing is done by software except for conversion between
NTSC video signals and bitmap digital images. The output image is updated
at a rate of 10 frames per second.

Recognition of objects can be naturally extended to the recognition of
situations.
Suppose that there is an ID code on every door in the building.
When the user stands in front of a door, the system detects where the user
is located and may understand what he/she intends to do by scanning the ID code
on the door and by processing the information related to it.

\subsection{Spoken Dialogue Processing}

The speech dialogue subsystem works as follows. First, a voice input
is acoustically analyzed by a built-in sound processing board.  Then,
a speech recognition module is invoked to output word sequences that
have been assigned higher scores by a probabilistic phoneme model and
phonetic dictionaries that are changed according to the situation.

These word sequences are syntactically and semantically
analyzed and disambiguated by applying a relatively loose grammar and a
restricted domain knowledge.  Using a semantic representation of the
input utterance, a plan recognition module extracts the speaker's
intention. For example, from the utterance ``I want to learn
computer science'' at a library front desk, the module interprets
the speaker's intention as ``The speaker wants to get information
about books on computer science (for example, the place
where he/she can get them).''

Once the system determines the speaker's intention, a response generation
module is invoked. This generates a response to satisfy the speaker's
request. Finally, the system's response is outputted as a voice by a voice
synthesis module.
This subsystem also sends a message to the visual message generator
about what graphical and/or textual information should be displayed
with the voice responses.

Speaker-independent continuous speech inputs are accepted without
special hardware. To obtain a high level of accuracy,
context-dependent phonetic hidden Markov models are used to construct
phoneme-level hypotheses \cite{Itou92}. The speech recognizer outputs
N-best word-level hypotheses.
As mentioned above, an appropriate phonetic dictionary is dynamically
selected by considering the speaker's real world situation.
Therefore, the perplexities or hypothetical spaces are always maintained in
tractable sizes without more advanced (and high-cost) speech technologies.

The semantic analyzer handles ambiguities in syntactic structures and
generates a semantic representation of the utterance. We applied a
preferential constraint satisfaction technique for disambiguation and
semantic analysis \cite{Nagao92a}.
For example, the following semantic representation is constructed from
the utterance ``I want to learn computer science'' at the library front desk.
\begin{verbatim}
(*want-1
;; *want-1 indicates that it is an instance
;; of frame *want.
   (:agent *i-1)
   (:theme (*learn-1
              (:agent *i-1)
              (:theme *computer-science-1)))
   (:situation *library-front-1)
   ;; :situation is added by the situation
   ;; awareness module.
)
\end{verbatim}

The plan recognition module determines the speaker's intention by
constructing his/her belief model and dynamically adjusting and expanding
the model as the conversation progresses \cite{Nagao93a}.
We use a plan library that is selected according to the situation.
In the case of the above example, {\tt library-front-plan} is selected.
Then, the recognized intention will be as follows\footnote{Actually,
the intention may have several candidates that are assigned numerical
preference values.}.
\begin{verbatim}
(*intend-to-know-1
;; *intend-to-know comes from (*want
;; (:theme *learn)) in library-front-plan.
   (:agent *speaker-1)
   ;; *i-1 is replaced with *speaker-1.
   (:theme (*location-of-bookshelf-1
              (:area *computer-science-1)))
   ;; *location-of-bookshelf is inserted by
   ;; means of plan inference.
)
\end{verbatim}

The spoken message generation module generates a response by using a
domain-dependent knowledge base and text templates (typical patterns
of utterances). It selects appropriate templates and combines them to
construct a response that satisfies the speaker's request.

\subsection{Integration of Linguistic and Non-Linguistic Contexts}

When the system detects a real world situation, it performs not only a
selection of
knowledge sources (e.g., phonetic/linguistic dictionaries) but also the
introduction of a non-linguistic context.
A non-linguistic context includes an object at which a user is
currently looking, a location where he/she currently is, graphical
information displayed on the screen, and chronological relations of
situation shifts. On the other hand, a linguistic context involves
semantic contents of utterances, displayed textual information, and
inferred beliefs and intentions (plans and goals) of the user.

Knowing the user's intention is necessary for natural human-computer
interaction. A real world situation is just a clue for it. However,
integrating the non-linguistic context introduced with the situation,
with the linguistic context constructed by dialogue processing, is an
important step.

In general, user's intentions are abductively inferred by using a plan
library \cite{Nagao93a}. A plan library is represented as an event network
whose nodes are events with their preconditions and effects, and links are
is-a/is-part-of relationships \cite{Kautz90}.

\begin{figure}[htbp]
\centering{\epsfile{file=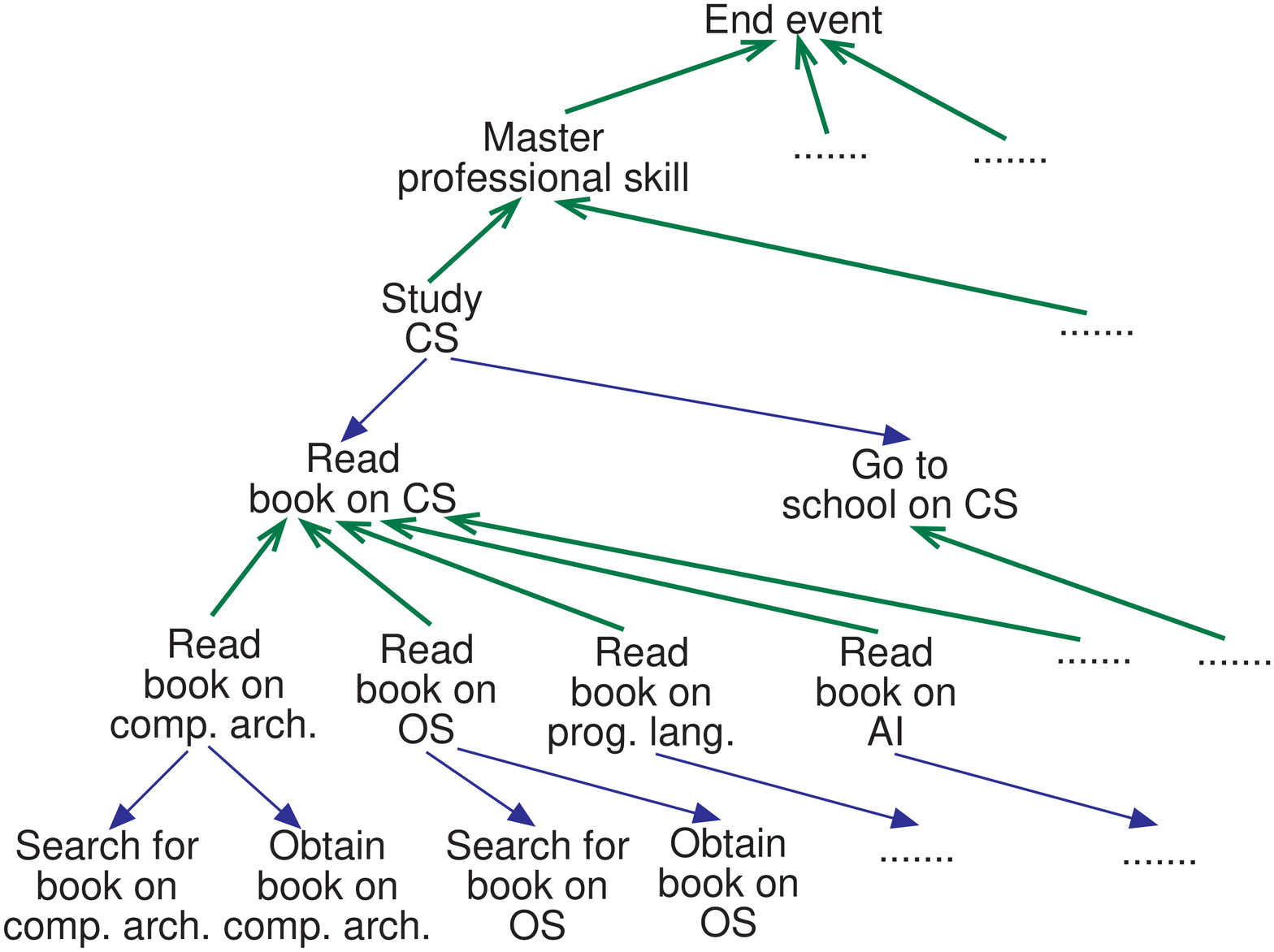,scale=.3}}
\caption{Plan Library {\tt comp-sci-bookshelf-plan}}
\label{plan-library}
\end{figure}

In the Ubiquitous Talker, plan inference is initially triggered by
introducing a new non-linguistic context, since the motivation of our
situated interaction is closely related to the physical actions for entering
a new situation.
For example, in a situation where a person is standing in front of a
bookshelf, for example a bookshelf on computer science, the situation will
motivate the person to search for a book on computer science, read it, and 
study it.
Therefore, when the dialogue system is aware of the situation by recognizing
the bookshelf's ID, the plan library shown in Figure \ref{plan-library} is
introduced and used for further plan inference. In this figure, the
upward-pointing thick arrows correspond to is-a (a-kind-of) relationships,
while downward-pointing thin arrows indicate has-a (part-of) relationships.

Introducing and focusing a specific plan library makes plan
recognition easier and more feasible.

Another connection between linguistic and non-linguistic contexts is
{\em deictic centers} \cite{Zancanaro93} that are possible referents
of deictic expressions. The object and the location in a
non-linguistic context can be current deictic centers. Also graphical
and textual information on the screen includes deictic centers.
Preferences on possible deictic centers as a referent are determined
based on coherence of a dialogue like the case of anaphora/ellipsis
resolution in a linguistic context \cite{Walker94}.

\section{Applications}

We now describe some sample applications of the Ubiquitous
Talker\footnote{Actually, the system accepts and speaks only in
Japanese. The example is a translation.}.
The current implementation of these applications is limited, so
the size of the vocabulary and the database are relatively small. However,
the concept presented here is sufficiently feasible and scaleable based on
current technology.
The examples show that situated interaction appropriately helps human
activities of information seeking, and supports human everyday life.

\subsection{Augmented Library}

The electronic library plan is to make all published material
computerized, and to construct an efficient, easy-to-use management and
retrieval system for them (i.e., a library in virtual space).
However, it is a very difficult task to digitize all published documents,
and as long as physical libraries exists in the real world, we must
think about the connection between physical and electronic libraries.
In the real world library, people must be able to retrieve useful
information from an electronic library by using the link between this and
physical libraries.

The Ubiquitous Talker can embody electronic information in
the confronted real world situation. Therefore in a physical library, when
the system is aware of the situation, it creates an information space
that augments the library.
This works well even if this information space is not yet mature and
well constructed. For example, this information does not need to
include the contents of books themselves, because books exist in the real
world.
We call the composite space of physical and electronic information an
{\em augmented library}.

When a user uses his/her Ubiquitous Talker at the library front desk 
(observing the signboard with the color-bar through it), a voice comes to 
the user from the system, saying ``This is the library of the Tokyo Institute
of Technology. Which area do you want?''. The user replies ``Computer
science''.  The system then shows a map and indicates the route to the
bookshelf for computer science books before saying ``Please take this
route.'' After he/she reaches the bookshelf, he/she sees it through the
system, which then says, (in fact the voice is generated by the system)
``Here we have books on computer science. What are you looking for?''
After replying ``A book on language,'' it then asks ``Which kind of language,
a programming language or natural language?'' Assuming the user does not
know what natural language is, he/she asks ``What is natural language?'' 
The response is ``Natural language is the language that humans use for
communication.''

When the user asks ``Where are the programming language books?,''
the system replies ``Books on programming languages are
on the third shelf of this bookshelf.'' He/she selects a book there and
looks at it through the system. Then the book seems to say ``The title
of this is `Object-oriented languages' and this was written by Mario
Tokoro'' and a description of the book is shown on the LCD display.
The user asks ``Tell me about the author,'' so it then displays and relates 
the profiles, achievements, and other publications of the author with
a photograph. After seeing the publication list, the user asks ``Where is 
the fourth book on this publication list?'' The answer is ``This is about 
computer architecture and is fifth from the right on the top shelf.''

In implementing the processing described above, we developed a
situation table which makes relationships between the identified situation,
resources for generating messages, and related phonetic/linguistic
dictionaries and knowledge bases (including plan libraries) for processing 
utterances. Part of the situation table is shown in Table \ref{relation}.

\begin{table*}[htbp]
\begin{center}
\caption{Relationships between Situations, Message Resources, and Dictionaries}
\label{relation}
\medskip
\begin{tabular}{|l|l|l|} \hline
{Situation} & {Resource of Messages} & {Dictionary/Knowledge Base} \\
\hline \hline
Library front & Area location guide & Dict1/Knowledge1 \\ \hline
................. & ................. & ................. \\ \hline
Bookshelf \#11 & Area/subarea description, & Dict11/Knowledge11 \\
`Computer science' & Subarea classification tree, & \\
 & Area/subarea location guide & \\ \hline
................. & ................. & ................. \\ \hline
Shelf \#113 & Subarea description, & Dict113/Knowledge113 \\
`Programming languages' & Subarea/book location guide & \\ \hline
................. & ................. & ................. \\ \hline
Book \#1135 & Book description, author database, & Dict1135/Knowledge1135 \\
`Object-oriented languages' & Book location guide & \\ \hline
................. & ................. & ................. \\ \hline
\end{tabular}
\end{center}
\end{table*}

When people know the exact name of a book that they want to
read and/or the name of its author before coming to the library, they
will immediately ask the location of the book at the library front desk.
However, it is not feasible for the system to accept speech inputs
of all names of the books that the library has and the names of their authors,
regardless of their areas. In this case, typing and/or hand-written
inputs must be accepted by the system.
In this paper, we concentrate on making speech interaction more
feasible by using the information about the physical environment.

\subsection{Talking Signboards}

Signboards are salient markers that act as representatives for certain
situations or objects. People see a signboard when they have some interests
in what is prepared beyond the signboard rather than the signboard itself.
The Ubiquitous Talker plays the role of a mediator for the
conversation between a human and a signboard.
In this case, in order to pinpoint a user's request, the system tries to make
the situation more specific by making suggestions on the display.

For example, suppose a man stands in front of a french restaurant, e.g.
`Maxim's de Paris,' and looks at its signboard through his Ubiquitous
Talker. The system identifies the restaurant from the ID code
attached to the signboard. Then, a voice message announces ``Welcome to
`Maxim's de Paris.' We are ready to tell you about the following
items,'' and a text message appears on the LCD display.
It describes ``1. Menu and Price, 2. Special Dishes recommended by
the Chef, 3. Wine List, ...'' When the user asks about the menu, the
system displays it and says ``Ok, here you are.'' The user
can also ask for a more detail description of any dish listed on the menu.

\section{Related Work}

There are several other researchers following similar directions to ours.
The major difference between our work and that of others is that we have
employed relatively light-weight and robust techniques for visual and language
processing and have combined them effectively. Below we discuss related
work in more detail.

\subsection{Ubiquitous Computing}

Ubiquitous computing \cite{Weiser93} proposes that very small computational
devices (i.e., ubiquitous computers) be embedded and integrated into
physical environments in such a way that they operate seamlessly
and almost transparently. These devices are aware of their physical
surroundings.
In contrast to ubiquitous computers, our barcode (color-code) system is a
low cost and reliable solution to making everything a computer.
Suppose that every page in a book has a unique barcode. When the user opens
a page, its page ID is detected by the system, so it can supply specific
information regarding the page. When the user adds some information to the
page, the system stores it with the page ID tagged for later retrieval.
This is almost the same as having a computer in every page of the book
without the cost. Our ID-aware system is better than ubiquitous computers
from the viewpoint of reliability and cost-performance, since it does not
require batteries and never breaks down.

\subsection{Chameleon: A Spatially-Aware Palmtop}

Chameleon \cite{Fitzmaurice93} is a spatially-aware palmtop computer.
It shows situated information according to its three-dimensional location
and orientation on a small LCD display.
This system is not as robust because it is aware of the situation only
from the location of the system itself. When physical objects change their
locations, it cannot recognize their movements.
The Ubiquitous Talker recognizes the situation from the visual
information of the real world. Therefore, it can follow the changes of the real
world. In addition, since our system integrates a speech dialogue technique,
the user can interact with the system more flexibly and deeply.

\subsection{Agents}

Our research is motivated by not only augmented reality but also
agent-oriented systems \cite{Laurel90,Nagao94b}. Agents (or robots)
recognize real world situations by using complicated perception techniques.
These techniques are currently not fully developed and difficult to use in
practice.
We have adopted an easy, reliable situation awareness technique by using
machine-recognizable IDs. This idea is also applicable for agents.

Another key issue of agent-oriented interfaces is the inference of user's
intention. Our solution is to use non-linguistic contexts for
focusing the user's attention and restricting the user's possible plans.
Real world situations can be information for selecting appropriate
plan libraries.
Therefore, our real world-oriented speech dialogue processing improves
not only the accuracy of speech/language analysis, but also the tractability
of intention recognition.

\section{Final Remarks}

We have developed an augmented reality system that integrates
situation awareness and spoken dialogue processing techniques. We
employed a barcode (color-code) system for easily recognizing real world
situations/objects.
Situation awareness contributes to reducing any hypothesis space
constructed during language processing, reducing the complexity
associated with understanding speech.
Situated conversation makes human-computer interaction more natural
and efficient by combining linguistic and non-linguistic contexts.

As a future research direction, we plan to integrate more communication
channels and modalities. For example, detection of the user's head/eye
orientation will be useful when it is necessary to determine the user's
more precise focus of attention in a complex situation where there are
several objects which are not placed in an ordered fashion.
In addition, we are interested in the integration of prosodic information
in speech recognition/synthesis.

We are also extending the technique for identifying real world situations.
Currently, our color-code system is so naive that it is difficult to
scale up. There are, however, some wireless electronic label systems
that use batteryless passive ICs. These technologies are applicable and would
improve our system. In addition, location awareness methods such as the
global positioning system are also useful for situation awareness, for example
Fitzmaurice's Chameleon system. Several real world information
(location, ID labels, time, distance, etc.) can increase the accuracy of
situation recognition and be used as sources of non-linguistic contexts.

\section*{Acknowledgments}

The authors would like to thank Mario Tokoro and colleagues at Sony CSL
for their encouragement and discussion.
Special thanks go to researchers in Speech Technology Group at Electrotechnical
Laboratory for their help to develop the speech recognition module.
We also extend our thanks to Hiroko Inui and Takashi Miyata
for their contributions to the implementation of the prototype system.


\end{document}